\documentclass{article}

\usepackage{graphicx}
\usepackage{multirow}
\usepackage{caption}
\usepackage{subcaption}

\usepackage[preprint]{neurips_2023}

\usepackage[utf8]{inputenc} 
\usepackage[T1]{fontenc}    
\usepackage{hyperref}       
\usepackage{url}            
\usepackage{booktabs}       
\usepackage{amsfonts}       
\usepackage{nicefrac}       
\usepackage{microtype}      
\usepackage{xcolor}         

\usepackage{amsmath,amsfonts,bm}

\def\eqref#1{equation~\ref{#1}}

\def\1{\bm{1}}

\DeclareMathAlphabet{\mathsfit}{\encodingdefault}{\sfdefault}{m}{sl}
\SetMathAlphabet{\mathsfit}{bold}{\encodingdefault}{\sfdefault}{bx}{n}

\title{LOWA: Localize Objects in the Wild with Attributes}

\author{
  Xiaoyuan Guo$^{*}$ \\
  Mineral \\
  \texttt{xiaoyuanguo@mineral.ai}
  \And
  Kezhen Chen \thanks{Xiaoyuan and Kezhen have equal contributions.} \\
  Mineral \\
  \texttt{kezhenchen@mineral.ai}
  \And
  Jinmeng Rao \\
  Mineral \\
  \texttt{jinmengrao@mineral.ai}
  \And
  Yawen Zhang \\
  Mineral \\
  \texttt{yawenz@mineral.ai}
  \And
  Baochen Sun \\
  Mineral \\
  \texttt{baochens@mineral.ai}
  \And
  Jie Yang \\
  Mineral \\
  \texttt{yangjie@mineral.ai}
}

\begin{document}
\maketitle

\begin{abstract}
We present LOWA, a novel method for localizing objects with attributes effectively in the wild. It aims to address the insufficiency of current open-vocabulary object detectors, which are limited by the lack of instance-level attribute classification and rare class names. To train LOWA, we propose a hybrid vision-language training strategy to learn object detection and recognition with class names as well as attribute information. With LOWA, users can not only detect objects with class names, but also able to localize objects by attributes. LOWA is built on top of a two-tower vision-language architecture and consists of a standard vision transformer as the image encoder and a similar transformer as the text encoder. To learn the alignment between visual and text inputs at the instance level, we train LOWA with three training steps: object-level training, attribute-aware learning, and free-text joint training of objects and attributes. This hybrid training strategy first ensures correct object detection, then incorporates instance-level attribute information, and finally balances the object class and attribute sensitivity. We evaluate our model performance of attribute classification and attribute localization on the Open-Vocabulary Attribute Detection (OVAD) benchmark and the Visual Attributes in the Wild (VAW) dataset, and experiments indicate strong zero-shot performance. Ablation studies additionally demonstrate the effectiveness of each training step of our approach.

\end{abstract}

\section{Introduction}
Recent advances in Vision-Language Pre-training (VLP)~\citep{radford2021learning} have led to significant progress in the Open-Vocabulary Object Detection (OVD) \citep{minderer2022simple, liu2023grounding, li2022glip}. Most of these models introduce language alignment for visual features of objects and have achieved impressive performance on zero/few-shot object detection with free-text inputs. Users can input natural language queries containing object names, and in response, the models localize objects that match these queries. Despite the remarkable progress, deploying these models in fine-grained domains remains challenging. Firstly, data bias constrains these models on fine-grained detection. In large-scale training data, general objects appear much more frequently than rare objects, and thus existing models usually have poor performance on text queries with terminologies. Secondly, using object names as detection queries is not always practical. In many real-world scenarios, users prefer to describe objects with their attributes (e.g., shape, color, texture) instead of precise terminologies. For example, using ``\textit{a black and white bird with red head feather}'' is more accessible in communication than using ``\textit{Antioquia Brushfinc}''. These issues present challenges for the development of OVD models.

\begin{figure*}[htp]
\centering
\includegraphics[width=0.8\textwidth]{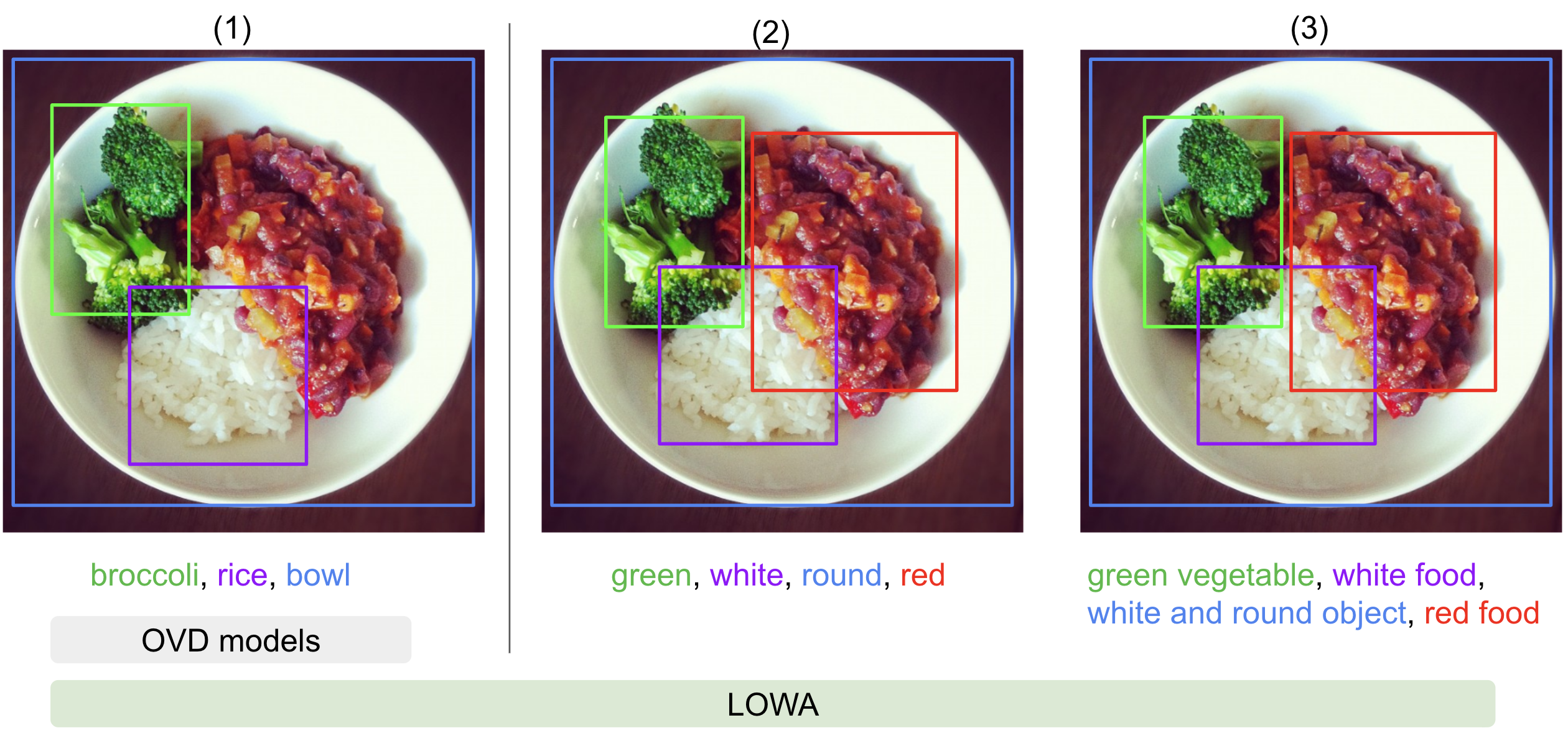}
\caption{Overview of LOWA functionality. OVD models take users' (1) object names as queries to predict; LOWA adapts more flexible queries including (1) object names, (2) attributes, and (3) compositions of object names and attributes.~\label{figure00}}
\end{figure*}

To address these issues, a promising direction is to use attributes as anchors to extend the generalization ability of existing OVD models. Objects share attributes, such as the color of \textit{red}, which can appear in both food and animal categories such as \textit{a red tomato} and \textit{a red bird}. Therefore, the knowledge of seen attributes can transfer to rare/unseen object categories by sharing the same object attributes or properties. Instead of only naming the object, using attributes to describe objects allows us to detect objects with more general free-text queries, and specify the unusual aspects of a familiar object (\textit{red food}, not just \textit{food}), see Fig.~\ref{figure00}. Furthermore, this method extends the ability to describe unfamiliar objects (\textit{hairy and four-legged}, not just \textit{object}) and to learn how to recognize new objects with few or no visual examples~\citep{farhadi2009describing}. To achieve this ability, models are required to disentangle attributes and object names. However, this research direction is still under-explored in existing OVD models.

We introduce our model LOWA, which extends OVD by localizing objects with attributes. Specifically, we focus on the attributes describing object appearance or status, such as color, shape, texture, pattern and action. Users can use both object classes and attribute descriptions to query the images and get detection results. The architecture of the model includes a standard vision transformer as the image encoder and another transformer as the text encoder. To enhance the attribute-awareness of the model, we design a three-step strategy for LOWA to align visual features with both object classes and attributes. In Step1, LOWA is trained to learn alignment between object classes and visual features and localize the objects. As a result, the model can detect general objects via text queries containing object names. In Step2, we incorporate the instance-level attributes to the model by training LOWA to disentangle object classes and attributes. In Step3, we concatenate attributes and objects together as free-text queries and use contrastive learning by randomly replacing either attributes or objects as negative samples. This step aims to enhance the model to align fine-grained text descriptions and visual features of objects.

To evaluate the attribute-awareness of the model, we tested LOWA on two attribute-annotated datasets, Visual Attributes in the Wild (VAW) \citep{pham2021learning} and Open-Vocabulary Attribute Detection (OVAD) \citep{bravo2022open}. Results show that LOWA outperforms all the open-vocabulary object detection baselines and achieves competitive performance compared to traditional two-stage attribute detection models. As a one-stage OVD model, LOWA has higher efficiency and flexibility. To further assess the attribute-aware object detection, we also evaluate on attribute localization with OVAD dataset. Results show that LOWA achieves state-of-the-art (SOTA) performance on the task compared to all the baselines. We believe that our model could provide a promising research direction in this field and benefits zero-shot/few-shot fine-grained object detection.
We summarize our contributions as follows:
\begin{itemize}
    \item We explore the under-represented attribute localization task based on open-vocabulary object detection and propose to learn instance attribute knowledge in open-vocabulary setting. Our model can perform open-vocabulary object detection with attribute classification all at once. To the best of our knowledge, this is the first open-vocabulary object detection model that is specially designed for learning fine-grained attributes. 
    \item We design a customized three-step training approach to train our model LOWA following a coarse-to-detailed learning schema, which enhances the ability to automatically focus on both object classes and attributes in free-style texts and localize the objects accurately.
    \item We evaluate our model on the attribute classification task and attribute localization task with two benchmarks to show the attribute awareness. On both tasks, our model outperforms all the baselines.
    \item We conduct ablation studies to show the effectiveness of our proposed hybrid training approach. Results indicate that each step in our training approach provides significant improvements.
\end{itemize}

\begin{figure*}[ht]
\centering
\includegraphics[width=0.8\textwidth]{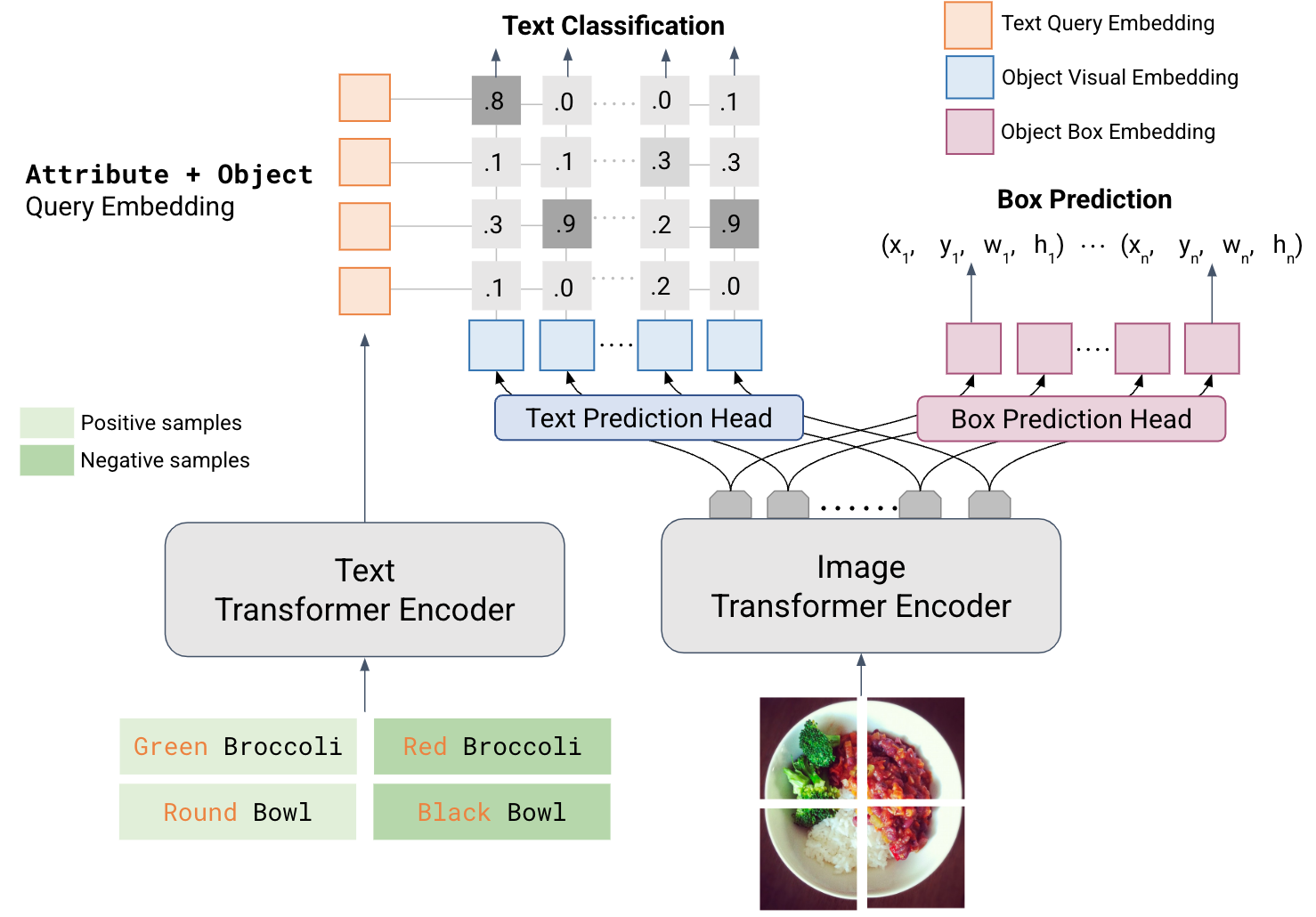}
\caption{Architecture Overview: The text encoder takes a set of text queries and generates a text embedding for each query. The image encoder takes the image patches as input and generate object-level embeddings. The object-level embeddings are passed to a text prediction head and a box prediction head. The text prediction head generates the object visual embeddings for object classification. The box prediction head generates the object box embeddings for box prediction.~\label{architecture-preview}}
\end{figure*}
 
\section{Method}
\label{sec:method}

The goal of LOWA is to localize target objects with free-text inputs, such as object names, object attributes or both. As shown in Fig.~\ref{figure00}, the model can localize objects even when the input is a combination of object names and attributes. Without explicit separation of attributes and object names, the model should be able to output good bounding boxes and correct matches with the given user's queries. To achieve this, we elaborate our method from the following aspects: model architecture we used, instance attribute learning process designed for LOWA and inference by users.

\subsection{Model Architecture}
LOWA uses a Vision Transformer Encoder as the image encoder and another Transformer Encoder as the text encoder, as depicted in Fig.~\ref{architecture-preview}. The text encoder generates a text embedding for each text query. Similarly as \citet{minderer2022simple}, we remove the token pooling and final projection layer from the text encoder. The image encoder encodes the image into a sequence of image embeddings, where each image embedding represents a detected object proposal. A text prediction head applies linear projection on each image embedding to generate an instance-level visual embedding for classification, and a box prediction head uses a Multi-Layer Perceptron (MLP) for bounding box regression. Thus, the total number of object proposals is equal to the length of image encoder inputs. We choose the architecture of ViT-L (768 embedding dimensions and 24 layers) with patch size 14 at input size 840 x 840 as the image encoder. Thus, the number of image embeddings is 3,600. Both the image encoder and text encoder are initialized from a pre-trained CLIP model \citep{radford2021learning}. As the model architecture only uses Transformer encoders, all the parameters take advantage of image-level pre-training in CLIP. The next section introduces the three-step training framework. 
\begin{figure*}[tp]
\centering
\includegraphics[width=0.9\textwidth]{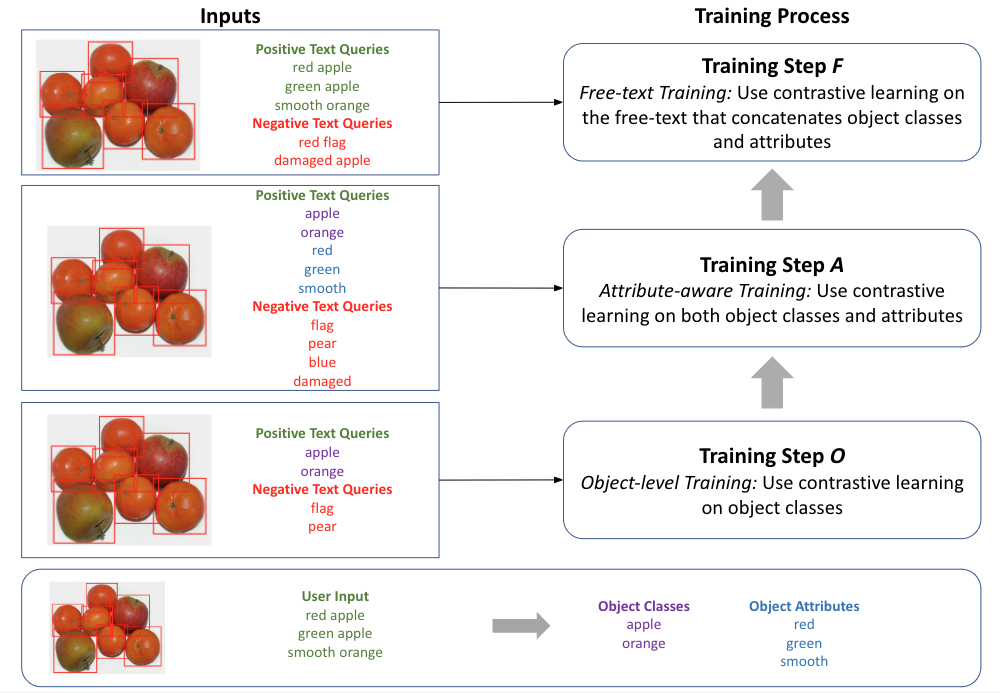}
\caption{Overview of our three-step training framework.  
~\label{pre-training}}
\end{figure*}
\subsection{Instance Attribute Learning}
Learning instance-level attribute knowledge is challenging for existing open-vocabulary object detectors with free-text queries as input, which requires the models to automatically identify the keywords without additional information. The free-text queries in our case can be any possible compositions of object names, attributes, object names + attributes, attributes + object names, etc. Without explicit separation of object names and attributes, existing object detection/attribute classification frameworks fail to meet such requirements.

Motivated by this, we propose a hybrid training strategy for LOWA shown in Fig.~\ref{pre-training} with a customized three-step learning process: \textbf{object-level training step} (Step$_{O}$ in short), \textbf{attribute-aware training step} (Step$_{A}$ in short) and \textbf{free-text training step} (Step$_{F}$ in short). This is to follow coarse-to-detailed alignment. Step$_{O}$ trains the model to detect objects by aligning object class names with visual representations. This helps to ensure that the correct objects are localized for Step$_{A}$ and thus the corresponding attribute details associated with a specific object can be learned. Step$_{F}$ leverages the previous object coarse and detailed knowledge to localize objects with free-form input queries. Simply mixing the attribute and object class information together and skipping Step$_{O}$ and Step$_{A}$ will affect the object localization accuracy and lead to misalignment of visual features and text features.

\textbf{Object-Level Training Step} 
At this step, the model is trained to align general object classes with visual objects. For each image, we provide the object class names as the positive text queries. A set of non-overlapped object class names is randomly picked from a candidate set as the negative text queries (We combine the object labels from two datasets, Objects365\citep{shao2019objects365} and LSA\citep{pham2022improving} as the object label set). The objective of this training step is to facilitate the matching of provided object classes with the corresponding object-level visual embeddings. Through ablation studies in Sec.~\ref{sec:ablation}, we have demonstrated that introducing coarse-grained object detection prior to fine-grained attributes yields a substantial improvement in accuracy for fine-grained object detection.

\textbf{Attribute-Aware Training Step}
The previous Step$_{O}$ has trained LOWA to learn the general open-vocabulary object detection at the coarse-grained level. In this step, we incorporate the attribute information into the model. As each object could have multiple attributes, we regard each attribute as an additional label to the object. The model is trained to align each visual embedding with multiple text queries. For example, \textit{``a red peeled apple''} corresponds to three positive text queries: \textit{``red''}, \textit{``peeled''}, and \textit{``apple''}. To create negative samples, the object classes and attributes are merged together to construct a label candidate set. For each object, we randomly sample some objects or attributes from the union set as the negative text queries. With these multi-label setting, the image encoder is optimized to disentangle the semantic meaning of attributes and classes.

\textbf{Free-Text Training Step}
With previous two steps, the model has awareness on both classes and attributes for each object. The third training step aims to enhance the vision-language alignment on more free-styled text queries containing both attributes and class names. For each object, we randomly pick one attribute from its attributes and concatenate it with the object class to construct a positive text query. For negative queries, the positive text queries are broken by either replace the object class or the attribute. We randomly picked either an object class or an attribute without overlaps. After this training step, LOWA learns to match text queries with fine-grained descriptions and instance-level visual embeddings. 

\textbf{Training Optimization}
All the training steps use the the same bipartite matching loss $\mathcal{L}_m$ introduced by DETR \citep{carion2020end} for object classification and bounding box prediction. We use the L1 loss $\mathcal{L}_1$ and the GIOU loss $\mathcal{L}_g$ for bounding box prediction. We also use the focal loss $\mathcal{L}_c$ as the classification loss for text labels, as each single object could correspond to multiple text labels. Instead of predicting the logits in a fixed global label space as closed-vocabulary detectors, the classification projection head outputs logits over the per-image label space defined by the queries. The matching loss $\mathcal{L}_m$ is formally defined as follows:
\begin{align} \label{eq:loss}
    \mathcal{L}_m = \mathcal{L}_1 + \mathcal{L}_g + \mathcal{L}_c
\end{align}

\subsection{Inference}~\label{sec:inference}
During inference, LOWA predicts the bounding boxes of objects in an image based on the provided text queries. Each text query could be a fine-grained description of an object with certain attributes. LOWA can be adapted to multiple tasks: (1) For \textbf{closed-vocabulary detection} task, all the classes will be used as the detection queries and pass to LOWA for detection; (2) For \textbf{attribute classification} on a targeted object, we first select the visual embedding that has the largest overlap between the predicted bounding box and the targeted object. Then, we use all the attribute candidates as text queries and compute the probabilities of all attributes on the selected visual embedding; (3) For \textbf{attribute localization}, instead of using the class names, we only use the object attributes as text queries to get the predicted bounding boxes. We use mean average recall to report the effectiveness regarding this task; (4) For \textbf{open-vocabulary object detection}, we match the free-text queries with all object bounding boxes and use a threshold (usually be 0.5) to find all detected objects.

\subsection{Implementation}
\label{sec:implementation}

\textbf{Training Data}
For Step$_{O}$, we uses Objects365\citep{shao2019objects365} and Visual Genome dataset\citep{krishna2017visual} for training. In this step, we only sample negative classes for constrastive learning. Step$_{A}$ and Step$_{F}$ are trained with four datasets: LSA\citep{pham2022improving}, PACO\citep{ramanathan2023paco}, Objects365 and MS COCO \citep{lin2014microsoft}. LSA and PACO have both object class and attribute annotations. Objects365 and MS COCO only have the object class labels. Therefore, we only use them to optimize object class learning. More training details can be found in appendix.

\textbf{Text Query Templates}
As the text query has both object classes and attributes, we cleaned the CLIP text templates by removing some templates containing adjectives such as \textit{``A photo of a large \{\}''} or \textit{``A photo of a black and white \{\}''}. These templates may confuse the model for attribute learning with irrelevant descriptive information. With manually inspection and cleaning, we finalize 60 templates for model training.

\textbf{Attribute Query Design}
Since our training data are federated from various sources, the attributes labeled are noisy and not necessarily descriptive vocabularies but phrases about object relationships. Despite this, we chose not to remove them during training because these attributes enhance the model to understand more free-styled text queries. However, concatenating these attributes and classes directly might cause inconsistency with users’ input habits. For example, people prefer to describe a \textit{cake} with the attribute \textit{on the table} as \textit{``cake on the table''} instead of \textit{``on the table cake''}. To avoid this issue, we train the model to adapt to both attribute+object and object+attribute orders. In this way, our model is capable of learning with the query \textit{``a photo of cake on the table''}.

\section{Experiments}

\textbf{Baselines}
Since our model is designed to localize objects in open-vocabulary environment, we choose two SOTA open-vocabulary detection models as the baselines: OWL-ViT \citep{minderer2022simple} and GroundingDINO \citep{liu2023grounding}. Both models have impressive open-vocabulary object detection ability and have shown attribute-awareness. 

\textbf{Evaluation Tasks}
As introduced in Sec.~\ref{sec:inference}, our model can perform multiple tasks but we focus on object localization with attributes. Therefore, we mainly evaluate our models on tasks: \textbf{attribute classification} and \textbf{attribute localization}. Attribute classification is the task of identifying the attributes of an object given its class name. The current attribute evaluation benchmarks are generally for this task. In contrast, attribute localization is the task of identifying the localization of an object given its attributes. Additionally, we report object detection performance on OVAD dataset following the previous works. 


\textbf{Datasets} We use two attribute-annotated datasets, VAW \citep{pham2021learning} and OVAD \citep{bravo2022open} for evaluation. VAW is constructed with VGPhraseCut \citep{wu2020vgcut} and GQA \citep{hudson2019gqa}. After cleaning, the dataset contains 620 attributes which describe object color, material, shape, size, texture and action. OVAD is an open-vocabulary attributes detection benchmark with clean and densely annotated objects and attributes (no training set is provided). Overall, the benchmark has 117 attribute classes for over 14,300 object instances.

\textbf{Evaluation Protocol}
Our model devotes to simplifying the dense object-attribute learning process and omitting the explicit object candidate generation. Therefore, our model localizes specific objects as well as predicts class names and attributes simultaneously. This is in contrast to many other open-vocabulary attribute classification models, which treat the task as two sub-tasks: object detection followed by category and attribute classification. We tested the two baselines under the \textbf{box-free} setting. Without target bounding boxes provided, we obtain predictions of bounding boxes and probabilities for a given image with a list of candidate attributes and object names.  Under the box-free setting, we first find the predicted box that has the largest Intersection over Union (IoU) with the ground truth box among all the predicted bounding boxes, and then use the corresponding object feature for attribute classification. We follow the same workflow for evaluating GroundingDINO and OWL-ViT. We calculate the mean average precision (mAP) over all the attributes as the metric to compare all the methods.

\textbf{Evaluation Metrics}
For attribute recognition, we follow VAW and OVAD to measure attribute prediction from a different perspective: mAP, mean average precision over all classes; mR@K, mean recall over all classes at top K predictions in each image; and F1@K, F1 scores over all classes at top K predictions in each image. K is selected as 8 for OVAD and 10 for VAW following the previous works' default settings. For attribute localization, we report mAR@10 results, which is the mean average recall at top 10 predictions.

\subsection{Attribute Recognition}
\textbf{OVAD Benchmark} We report the overall attribute recognition performance in Tab.~\ref{table:ovad_results} using the metrics mAP, mR@15 and F1@15. As the occurrence of attributes is long-tailed, we split the attributes into Head, Medium and Tail and present the model mAP performance for each of them. This helps us to inspect model's generalization ability in handling common attributes and rare attributes. Additionally, we evaluate mAP at 0.5 IoU for the open-vocabulary object detection task on the 80 class object set, called OVD-80. 
As we can see, our model LOWA exhibits the best performance overall and is able to recognize attributes with different frequencies, especially the rare attributes. In contrast, GroundingDINO and OWLViT are not good at attribute classification, despite their decent performance on object detection task, as shown in the OVD-80 results. For a fair comparison, we also continued training OWL-ViT for 100K steps using the same training data as described in Sec.~\ref{sec:implementation}. We report this model's performance with the model as OWL-ViT (ViTL/14, Cont.).  However, the attribute classification ability drops compared to the original OWL-ViT model.

We also put the comparison results between our model with recent traditional attribute classification models in Tab.~\ref{table:ovad_others}. Most of the models in this table lack the object detection components, and thus they are evaluated under a simpler \textbf{box-given} setting. The ground truth locations of objects are provided and the models only perform attribute classification on each given object. Based on the results, our model achieves competitive results even compared with these traditional attribute classification models. The OvarNet is also a box-free model designed for open-vocabulary attribute classification and achieves impressive performance. However, OvarNet is trained via a teacher-student process to learn attribute detection. Our model uses a one-stage detection process, which is more efficient and general to adapt to multiple visual tasks. Also, our model significantly outperforms OvarNet on object detection results (improved by 24.5\% on all objects). Results from Tab.~\ref{table:ovad_results} and Tab.~\ref{table:ovad_others} provide promising evidence to show that our model has better attribute-awareness compared to other OVD baselines while keeping strong object detection ability.
\begin{table}[tb]
\caption{Attribute recognition and object detection results on OVAD benchmark.~\label{table:ovad_results} (Bold indicates the optimal performance.)}
\centering
\resizebox{\textwidth}{!}{
\begin{tabular}{l|cccccc|ccc}
\hline
\multirow{2}{*}{\textbf{Method}} & \multicolumn{6}{c|}{\textbf{OVAD}} & \multicolumn{3}{c}{\textbf{Generalized OVD-80 (AP$_{50}$)}} \\ \cline{2-10} 
 & \textbf{mAP(all)} & \textbf{mR@8} & \multicolumn{1}{l|}{\textbf{F1@8}} & \textbf{Head} & \textbf{Medium} & \textbf{Tail} & \textbf{Novel$_{(32)}$} & \textbf{Base$_{(48)}$} & \textbf{All$_{(80)}$} \\ \hline
GroundingDINO (Swin-T) & 8.9 & 7.8 & \multicolumn{1}{l|}{8.1} & 34.4 & 8.1 & 0.8 & 56.1 & 54.8 & 55.3 \\
GroundingDINO (Swin-B)& 8.9 & 8.0 & \multicolumn{1}{l|}{9.0} & 36.0 & 7.6 & 0.7 & \textbf{70.5} & 64.6 & 67.0 \\
OWL-ViT (ViTL/14) & 11.1 & 13.9 & \multicolumn{1}{l|}{15.8} & 42.8 & 10.0 & 1.3 & 58.6 & 56.4 & 57.3 \\
OWL-ViT (ViTL/14, Cont.) & 10.2 & 12.3 & \multicolumn{1}{l|}{14.0} & 40.3 & 9.0 & 0.9 & 69.3 & \textbf{67.4} & \textbf{68.2} \\ \hline
\textbf{LOWA (Ours)} & \textbf{18.7} & \textbf{35.3} &  \multicolumn{1}{l|}{\textbf{39.4}} & \textbf{58.0} & \textbf{20.4} & \textbf{2.6} & 68.2  & 67.0  & 67.5 \\ \hline
\end{tabular}
}
\end{table}

\begin{table}[tb]
\caption{Performance comparison with attribute classification models on OVAD benchmark.~\label{table:ovad_others}}
\centering
\resizebox{\textwidth}{!}{
\begin{tabular}{l|l|cllccc|ccc}
\hline
\multirow{2}{*}{\textbf{Method}} & \multirow{2}{*}{\textbf{Box}} & \multicolumn{6}{c|}{\textbf{OVAD}} & \multicolumn{3}{c}{\textbf{Generalized OVD-80 (AP$_{50}$)}} \\ \cline{3-11} 
 &  & \textbf{mAP(all)} & \textbf{mR@8} & \multicolumn{1}{l|}{\textbf{F1@8}} & \textbf{Head} & \textbf{Medium} & \textbf{Tail} & \textbf{Novel$_{(32)}$} & \textbf{Base$_{(48)}$} & \textbf{All$_{(80)}$} \\ \hline
OV-Faster-RCNN~\citep{bravo2022open} & given  & 11.7 & -- & \multicolumn{1}{l|}{--} & 34.4 & 13.1 & 1.9 & 0.3 & 53.3 & 32.1 \\
VL-PLM~\citep{zhao2022vlplm} & given & 13.2 & -- & \multicolumn{1}{l|}{--} & 32.6 & 16.3 & 2.6 & 19.7 & 58.8 & 43.2 \\
Detic~\citep{zhou2022detic} & given & 13.3 & -- & \multicolumn{1}{l|}{--} & 44.4 & 13.4 & 2.3 & 20.0 & 49.2 & 37.5 \\
LocOv~\citep{zhou2022locov} & given & 14.9 & -- & \multicolumn{1}{l|}{--} & 42.8 & 17.2 & 2.2 & 22.5 & 52.5 & 40.5 \\
OVR~\citep{zareian2022ovr} & given & 15.1 & -- & \multicolumn{1}{l|}{--} & 46.3 & 16.7 & 2.1 & 17.9 & 51.8 & 38.2 \\
OVAD~\citep{bravo2022open} & given & 18.8 & -- & \multicolumn{1}{l|}{--} & 47.7 & 22.0 & 4.6 & 24.7 & 49.1 & 39.3 \\ \hline
OvarNet(ViT-B16)~\citep{chen2023ovarnet} & free & \textbf{27.2} & -- & \multicolumn{1}{l|}{--} & 56.8 & \textbf{33.6} & \textbf{8.9} & 60.4 & 35.2 & 54.2 \\ \hline
\textbf{LOWA (Ours)}  & free & 18.7 & \textbf{35.3} & \multicolumn{1}{l|}{\textbf{39.4}} & \textbf{58.0} & 20.4 & 2.6 & \textbf{68.2}  & \textbf{67.0}  & \textbf{67.5}  \\ \hline
\end{tabular}
}
\end{table}

\textbf{VAW Benchmark} Similar to the OVAD dataset, we evaluate models using zero-shot attribute classification task on the VAW validation set under a box-free setting. Table~\ref{table:vaw-result} shows the performance comparison between all the baseline models and our model. Based on the results, our model outperforms all the baseline models on the attribute classification task and thus shows enhanced attribute-awareness. Table~\ref{table:vaw-attr} also reports the performance on each attribute category in the VAW dataset. Results show that our proposed training framework improves almost all the attribute categories. Color, material, and action have the most significant improvements. One possibility is that color and material are the most common attributes across most of the objects. Training over large-scale data improves the model generalization on common attributes. On the other hand, attributes in action category are highly correlated to object types. Training model on large-scaled data may reduce the data bias on unbalance attributes. Thus, these observations provide additional evidence on the attribute-awareness and generalization of our model.

\begin{table}[t]
\centering
\caption{Attribute recognition results on VAW validation set.~\label{table:vaw-result} (Bold indicates the best performance.)}
\resizebox{0.8\textwidth}{!}{
\begin{tabular}{l|ccc|ccc}
\hline
\textbf{Model} & \textbf{mAP(all)} & \textbf{mR@15} & \textbf{F1@15} & \textbf{Head} & \textbf{Medium} & \textbf{Tail} \\ \hline
GroundingDINO (Swin-B)  & 30.2                & 2.0           & 3.7          & 33.8            & 28.7              & 20.8  \\
GroundingDINO (Swin-T)  & 30.0                & 2.7           & 5.0          & 33.9            & 28.4              & 20.2  \\
OWL-ViT (ViTL/14)        & 37.4              & 12.3           & 20.4         & 41.8            & 35.9              & 25.3   \\
OWL-ViT (ViTL/14, Cont.)         & 36.1         & 9.4             & 16.1          & 40.3            & 34.4              & 25.6  \\ \hline
\textbf{LOWA (Ours)}           & \textbf{42.6}              & \textbf{36.9}           & \textbf{45.4}          & \textbf{46.4}          & \textbf{41.0}            & \textbf{32.9}       \\ \hline
\end{tabular}
}
\end{table}

\subsection{Attribute Localization}
We further evaluate the effectiveness of our model on attribute localization. To this end, we reuse the OVAD dataset and only use attributes as queries to localize the target objects such as \textit{``red''} or \textit{``smooth''} object. Specifically, we evaluate the attribute awareness for each attribute category. Table~\ref{ovad-ar10} reports nine representative categories, with more reported in appendix. From the table, LOWA significantly outperforms all the baselines, which provides further evidence for the attribute-awareness.

\begin{table}[t]
\centering
\caption{Attribute recognition performance (mAP) on each attribute category of VAW dataset.~\label{table:vaw-attr} (Bold indicates the best performance.)}
\resizebox{0.8\textwidth}{!}{
\begin{tabular}{l|ccccccc}
\hline
\textbf{Model} & \textbf{Color} & \textbf{Material} & \textbf{Shape} & \textbf{Size} & \textbf{Texture} & \textbf{Action} & \textbf{Other} \\ \hline
GroundingDINO (Swin-B)  & 20.9             & 20.3           & 29.1            & 36.2            & 32.5               & 15.6          & 36.3  \\
GroundingDINO (Swin-T)  & 20.6             & 20.8           & 31.1            & 36.3            & 32.5               & 14.7          & 36.1  \\
OWL-ViT (ViTL/14)         & 29.8           & 31.6           & 39.7            & 43.8            & 38.3               & 24.7          & 42.2  \\
OWL-ViT (ViTL/14, Cont.)        & 25.9             & 29.4                & 37.0             & 42.8           &   38.2           & 23.4             & 41.8  \\ \hline
\textbf{LOWA (Ours)}           & \textbf{37.9}           & \textbf{37.5}              & \textbf{41.2}          & \textbf{47.9}          & \textbf{42.7}             & \textbf{34.1}            & \textbf{46.1}            \\ \hline
\end{tabular}
}
\end{table}

\begin{table}[t]
\caption{Performance (AR@10) of Attribute Localization on OVAD dataset}~\label{ovad-ar10}
\resizebox{\textwidth}{!}{
\begin{tabular}{l|ccccccccc}
\hline
\textbf{Model} & \textbf{Cleanliness} & \textbf{Pattern} & \textbf{State} & \textbf{Type} & \textbf{Material} & \textbf{Size} & \textbf{Texture} & \textbf{Length} & \textbf{Tone} \\ \hline
\begin{tabular}[c]{@{}l@{}}OWL-ViT \end{tabular} & 20.1 & 24.4 & 19.9 & 12.8 & 29.7 & 23.8 & 12.9 & 10.8 & 8.8 \\ 
\begin{tabular}[c]{@{}l@{}}OWL-ViT (ViTL/14. Cont.) \end{tabular} & \textbf{25.0} & 37.3 & 21.5 & 15.4 & 28.6 & 22.3 & 16.4 & 20.2 & 10.7 \\ \hline
\textbf{LOWA (Ours)} & 23.3 & \textbf{37.9} & \textbf{21.8} & \textbf{29.6} & \textbf{32.2} & \textbf{26.6} & \textbf{22.5} & \textbf{27.4} & \textbf{15.3} \\ \hline
\end{tabular}
}
\end{table}

\section{Ablation Study}
\label{sec:ablation}

To investigate the effectiveness of our training strategy, we conduct ablation studies by removing each training step. We increase the training steps in each ablation study to keep the same number of total training steps for fair comparison. We increase the Step$_{O}$ or Step$_{F}$ to 200K steps for setting Step$_{O}$+$_{F}$ and Step$_{A}$+$_{F}$. By removing Step$_{F}$, we increase 25K steps for each previous step. Furthermore, as we mentioned in Section ~\ref{sec:method}, the goal of Step$_{F}$ is to enhance vision and language matching instead of improving attribute and object detection. We also investigate whether Step$_{F}$ follows our assumption by increasing the training steps in Step$_{F}$ to 150K steps. Table~\ref{table:ablation} shows the performance comparison of all the studies. Results show that the model trained with three steps outperforms all the other designs that miss one training step. These studies convince that our training framework is effective to enhance the fine-grained object detection of OVD  models. Training the model to learn open-vocabulary object detection from coarse-grained to fine-grained features is necessary to disentangle classes and attributes. Also, we prove our assumption that longer Step$_{F}$ does not help attribute and object detection.

\begin{table}[t]
\caption{Ablation studies on OVAD benchmark.~\label{table:ablation}}
\centering
\resizebox{\textwidth}{!}{
\begin{tabular}{l|c|cll|ccc}
\hline
\multirow{2}{*}{\textbf{Method}} & \multirow{2}{*}{\textbf{Training Steps}} & \multicolumn{3}{c|}{\textbf{OVAD}} & \multicolumn{3}{c}{\textbf{Generalized OVD-80 (AP$_{50}$)}} \\ \cline{3-8} 
 &  & \textbf{mAP(all)} & \textbf{mR@8} & \multicolumn{1}{l|}{\textbf{F1@8}} & \textbf{Novel$_{(32)}$} & \textbf{Base$_{(48)}$} & \textbf{All$_{(80)}$} \\ \hline
Step$_{O}$+$_{A}$ (Ours) & 250K & 17.0 & 27.7 & \multicolumn{1}{l|}{31.7} & \textbf{75.4} & \textbf{71.8} & \textbf{73.3} \\
Step$_{A}$+$_{F}$ (Ours)& 250K & 16.3 & 27.3 & \multicolumn{1}{l|}{30.5} & 45.9 & 40.1 & 42.4 \\
Step$_{O}$+$_{F}$ (Ours)& 250K & 18.1 & 32.4 & \multicolumn{1}{l|}{36.3} &66.4  &66.1  &72.1  \\ \hline
{Step$_{O}$+$_{A}$+$_{F}$} (Ours)& 350K &  18.5 & 34.4 &  \multicolumn{1}{l|}{38.5} & 70.2  & 68.7  & 69.3 \\ \hline
{Step$_{O}$+$_{A}$+$_{F}$} (Ours)& 250K & \textbf{18.7} & \textbf{35.3} &  \multicolumn{1}{l|}{\textbf{39.4}} & 68.2  & 67.0  & 67.5 \\ \hline
\end{tabular}
}
\end{table}


\section{Related Work}

\textbf{Attribute Detection}
Attribute detection is important for domain-specific detection tasks. However, attribute detection at instance-level is still under-explored because of some difficulties. The attribute definition varies across different tasks and constructing the datasets requires expensive costs to annotate both object classes and attributes at instance-level. Even though, many researchers have made great efforts towards collecting clean and high-quality datasets and they are generally classified into four categories: patch-labeled, partially-labeled, sparsely-labeled and densely-labeled. Patch-labeled means collecting a pair of attribute and object labels and cropping the instance from the original images \citep{isola2015discovering}; partially-labeled datasets are often with one or two instances annotated with bounding boxes, class names and attribute information \citep{pham2021learning};  sparsely-labeled datasets collect attribute information from image captions by extraction of adjective vocabularies or phrases \citep{krishna2017visual, pham2022improving}; densely-labeled datasets have both object and attribute labels for each object in an image \citep{bravo2022open}. 
Many researchers also proposed and developed algorithms for attribute detection. For example, OADis \citep{saini2022disentangling} detects object attributes by disentangling visual embeddings for attributes and objects using a CNN-based method with evaluations. SCoNE \citep{pham2021learning} aims to predict visual attributes in the wild and formulates the attribute detection as a multi-class classification problem with instance binary mask as supervision. \citet{bravo2022open} designs a model that includes a frozen language model and an object detector based on Faster-RCNN. Extracted nouns and noun phrases from image caption datasets are used to train the model for attribute detection. OpenTAP~\cite{pham2022improving} considers attributes as any phrases with semantic properties and relationships, and the authors use a pre-trained ResNet50 as the visual feature encoder and a Transformer model to fuse visual features, bounding boxes, and text embeddings. Although these works achieve remarkable results, they perform attribute classification on ground-truth object locations instead of detecting objects and attributes. Our model aims to localize objects with specific attributes in the open-vocabulary setting, which is more general in real-world scenarios. OvarNet \citep{chen2023ovarnet} focuses on the open-vocabulary detection on objects and attributes with a two-stage approach. A detection model with the architecture of Faster-RCNN \citep{ren2015faster} is used to acquire the object candidate proposals first and then a CLIP model \citep{radford2021learning} finetuned with attribute data performs attribute prediction on each detected object. While this model achieves great performance on attribute and object detection, it uses a two-stage detection process. Our model uses a one-stage process to directly align image features and text queries, which improves flexibility and efficiency.

\textbf{Open-Vocabulary Object Detection}
Open-vocabulary object detection (OVD) aims to detect target objects that not present in the training/base class vocabulary \citep{rasheed2022bridging}. For example, ViLD \citep{gu2021open} performs open-vocabulary object detection via vision and language knowledge distillation. It distills knowledge from a pre-trained open-vocabulary image classification model into a two-stage detector. MEDet \citep{chen2022open} introduces a proposal mining and prediction equalization framework to alleviate the proposal-level vision-language alignment and base-novel category prediction balance. It refines the inherited vision-semantic knowledge in a coarse-to-fine and online manner to allow detection-oriented proposal-level feature alignment. Meanwhile, it equalizes prediction confidence by reinforcing novel category predictions with an offline class-wise adjustment. DetPro \citep{du2022learning} proposes to learn continuous prompt representation for OVD. \citet{rasheed2022bridging} proposed region-based knowledge distillation to adapt image-centred CLIP embeddings for local regions, aiming to improve alignment between region and language embeddings. GroundingDINO \citep{liu2023grounding} is an open-set object detector that can detect arbitrary objects with human inputs, such as category names or referring expressions. It fuses language and vision modalities and adopts a DETR-like structure to directly predict object categories and bounding boxes. Although these models have made significant progress, they either adopt a two-stage detector or perform vision and language fusion. On the contrary, our model uses a one-stage detection process and align vision and textual features via contrastive learning, which provides more flexibility and efficiency. The most similar model compared to our model is OWL-ViT \citep{minderer2022simple}. OWL-ViT also uses contrastive learning to train two encoders to align visual features and text queries. However, we propose the novel three-step training framework that aligns both objects and attributes with visual features. OWL-ViT is only trained to align object classes with visual features, which lacks attribute-awareness.

\section{Conclusion}
In this work, we introduce LOWA, a new open-vocabulary object detection model with fine-grained attribute-awareness. We also present a novel training framework to incorporate attribute information, which is fully compatible with existing OVD models. Our model trained on large-scale data can perform open-vocabulary object detection with free-text inputs in a one-stage manner. We also compare our model with several SOTA open-vocabulary detection models and observe a significant improvement on two popular benchmarks. Results show that LOWA has strong zero-shot detection performance and can disentangle object classes and attributes. Ablation studies indicate that each step of our training framework is effective and important. We believe that this work could improve the bottleneck of existing OVD models on fine-grained domain adaption. The novel ideas and observations could provide valuable insights to researchers in this field.

In the future, we plan to explore two research directions. Firstly, object detection based on multi-modality queries, such as images and documents, could be a very interesting future work. Secondly, feature alignment between vision models and conversational large language models is a promising approach. Performing fine-grained open-vocabulary object detection via conversations could be an impressive direction to explore.

{
\small


\bibliographystyle{lowa_arXiv}
\bibliography{lowa_arXiv}

}

\appendix
\section*{Appendix}
\section{Implementation Details}

\textbf{One Attribute at a Time}
To keep consistent with users' input style, we concatenate attributes and objects together so that our model can identify object class and attribute automatically with self-attention. We suggest that the text query should contain only one attribute and one object name, although we do not put any constraints on users' text queries. This is also consistent with our performance evaluation, which inspects each single attribute detection ability of our model.

\textbf{Negative Attribute Augmentation}
We also supplement necessary negative attributes with a large language model (LLM) for more concise and efficient attribute pattern learning. For example, the attribute \textit{``sliced''} describes the state of an object. Therefore, a random negative attribute such as \textit{``black''} will not help the model learn the attribute \textit{``sliced''}. However, the negative attribute \textit{``whole''} will teach the model to pay attention to the state attribute.

\textbf{Model Training Details}
\label{sec:training_details}
We use 8x8 slices TPU v3 with a batch size of 128 for training. We set 100K steps at the first training step, 100K steps at the second training step, and 50K steps at the third training step. The learning rate for the image encoder is 1e-5, and the learning rate for the text encoder is 2e-6.

\section{Attribute Learning Analysis}
\begin{figure*}[htp]
\centering
\includegraphics[width=\textwidth]{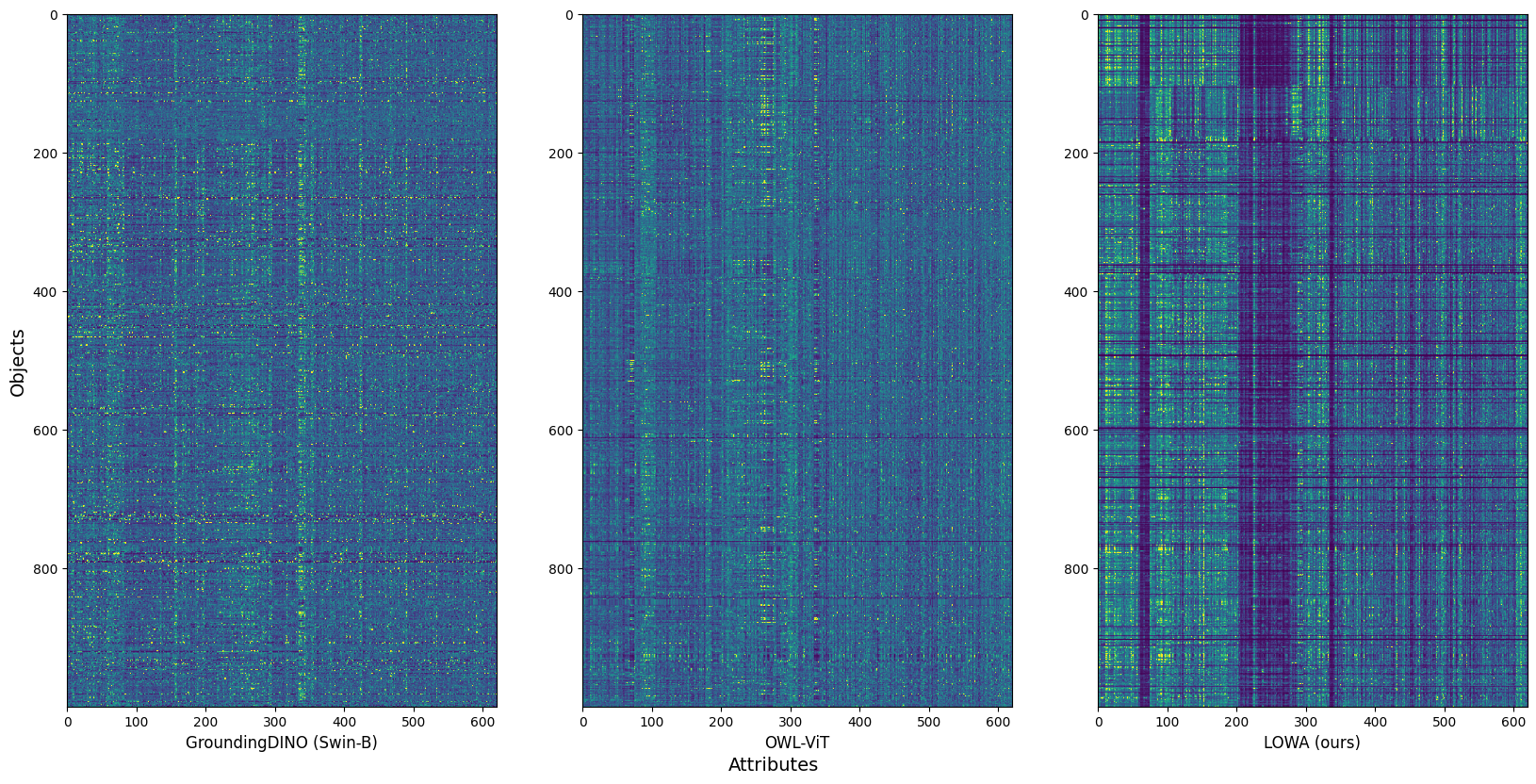}
\caption{Visual comparison of the normalized logits of each object+attribute query on VAW dataset predicted by GroundingDINO (Swin-B, left), OWL-ViT (middle), and LOWA (ours, right).~\label{logits_matrices}}
\end{figure*}

To further exemplify the attribute-awareness of our method, we provide a visual comparison of attribute-level logits matrices predicted by LOWA and other benchmark methods. The visual comparison of logits matrices for object+attribute queries on the VAW dataset is presented in Fig.~\ref{logits_matrices}, highlighting GroundingDINO (Swin-B, on the left), OWL-ViT (center), and LOWA (our method, on the right). All logits matrices are normalized along the X-axis, with only the first 1,000 objects from the VAW dataset displayed due to space constraints. Evidently, the logits matrix for LOWA presents significant attribute-varied distribution patterns, whereas the matrices for GroundingDINO and OWL-ViT demonstrate comparatively random patterns. This suggests that LOWA exhibits a higher degree of attribute awareness compared to the other benchmark methods.

\section{Additional Experimental Results}
In Tab.~\ref{table:vaw-attr2}, we provide additional information about LOWA's attribute recognition performance on each attribute category of the OVAD dataset. We use the metric mAP to measure the performance. The results show that our model can achieve good performance for category-wise attribute recognition as well.   
\begin{table}[htp]
\centering
\caption{Attribute recognition performance (mAP) on each attribute category of OVAD dataset.~\label{table:vaw-attr2} (We truncate the table and continue the rest. Bold indicates the best performance.)}
\resizebox{\textwidth}{!}{
\begin{tabular}{l|ccccccc}
\hline
\textbf{Model} & \textbf{Cleanliness} & \textbf{Color} & \textbf{Pattern} & \textbf{Color Quantity} & \textbf{Cooked} & \textbf{Face Expression} & \textbf{Gender}\\ \hline
GroundingDINO (Swin-B)  & 16.0             & 4.5          & 4.9           & 23.3            & 6.1               & 2.0          & 9.3    \\
GroundingDINO (Swin-T)  & 13.9             & 4.9           & 4.8            & 22.0            & 5.8              & 2.4          & 12.6  \\
OWL-ViT (ViTL/14)         & 26.6           & 6.8          & 6.0          & 27.0           & 5.4              & 1.2          & 12.5  \\
OWL-ViT (ViTL/14, Cont.)        & 15.5             & 6.1                & 6.1            & 25.3           &   4.4           & 1.6            & 10.2  \\ \hline
\textbf{LOWA (Ours)}           & \textbf{32.8}           & \textbf{9.7}              & \textbf{10.2}          & \textbf{31.3}          & \textbf{41.0}             & \textbf{8.3}            & \textbf{16.5}         \\ \hline
\end{tabular}
}
\resizebox{\textwidth}{!}{
\begin{tabular}{cccccccccccc}
\hline
 \textbf{Group} &\textbf{Length} & \textbf{Tone} & \textbf{Type} & \textbf{Material} & \textbf{Maturity} & \textbf{Optical Property} & \textbf{Order} & \textbf{Position} & \textbf{Size} &\textbf{State} &\textbf{Texture}\\ \hline
50.0  & 6.6 & 20.6            & 6.2           & 5.4            & 18.7            & 18.8               & 9.0         & 28.7   & 19.8 & 5.0 & 23.3\\
50.1  & 7.3 & 21.1            & \textbf{9.2}           & 5.2           & 21.0            & 17.2              & 8.7         & 29.1   & 16.9 & 4.5 & 21.3\\
50.5  & 10.5 & 25.2          & 4.8          & 9.8            & 16.3           & 22.3               & 15.6          & 31.4   & 20.1 & 5.8 & 27.5\\
50.5 & 9.8 & 23.6           & 5.8                & 8.2            & 16.5          &   18.8           & 8.1            & 32.1  & 25.5 & 4.9 & 26.2\\ \hline
\textbf{51.1}  &\textbf{11.9} & \textbf{29.8}           & 6.8              & \textbf{24.9}          & \textbf{45.6}          & \textbf{38.0}             & \textbf{16.4}            & \textbf{44.9}        &\textbf{51.3}  &\textbf{15.4}   &\textbf{45.2} \\ \hline
\end{tabular}
}
\end{table}

We supplement the left results of attribute localization on the OVAD dataset with the metric AR@10 in Tab.~\ref{ovad-ar10-2}. Generally, LOWA exhibits better performance than others regarding the attribute categories \textit{Cooked}, \textit{Order} and \textit{Optical Property}. However, it shows lower performance than others with attributes like \textit{Color}, \textit{Face Expression}, \textit{Gender}, and so on. One possible reason is the attribute bias with certain object categories. For example,  only \textit{person} objects are annotated with \textit{Color} information in the OVAD dataset. However, using $Color$-related attributes to localize \textit{person} is challenging as other objects may show more relevance. Instead, our model LOWA is able to detect various objects with $Color$-related attributes (See Sec.~\ref{sec:vis} for visualization).    


\begin{table}[t]
\caption{More performance (AR@10) of Attribute Localization on OVAD dataset}~\label{ovad-ar10-2}
\resizebox{\textwidth}{!}{
\begin{tabular}{l|ccc|ccccccc}
\hline
\textbf{Model} & \textbf{Cooked} & \textbf{Order}& \begin{tabular}[c]{@{}l@{}}\textbf{Optical} \\ \textbf{Property}\end{tabular}  & \textbf{Color} & \begin{tabular}[c]{@{}l@{}}\textbf{Color} \\ \textbf{Quantity}\end{tabular}  & \begin{tabular}[c]{@{}l@{}} \textbf{Face} \\ \textbf{Expression}\end{tabular} & \textbf{Gender} & \textbf{Group} & \textbf{Maturity}  & \textbf{Position} \\ \hline
\begin{tabular}[c]{@{}l@{}}OWL-ViT \end{tabular} & 30.2 & 19.8 &19.2 & 0.5 & \textbf{0.8} & \textbf{60.1} & \textbf{72.1} & \textbf{25.3} & \textbf{60.6}   & 21.0\\ 
\begin{tabular}[c]{@{}l@{}}OWL-ViT (ViTL/14. Cont.) \end{tabular} & 22.4  & 20.2  & 20.3 & \textbf{1.6} & 0.7  & 47.7 & 59.2 & 19.2 & 44.5 & \textbf{21.5}\\ \hline
\textbf{LOWA (Ours)}& \textbf{31.8} & \textbf{22.9} & \textbf{23.5}  & 1.0 & 0.7  & 47.4 & 58.3 & 21.0 & 50.9  & 17.7\\ \hline
\end{tabular}
}
\end{table}

\section{Example Visualizations}~\label{sec:vis}
\begin{figure*}[tp]
\centering
\includegraphics[width=\textwidth]{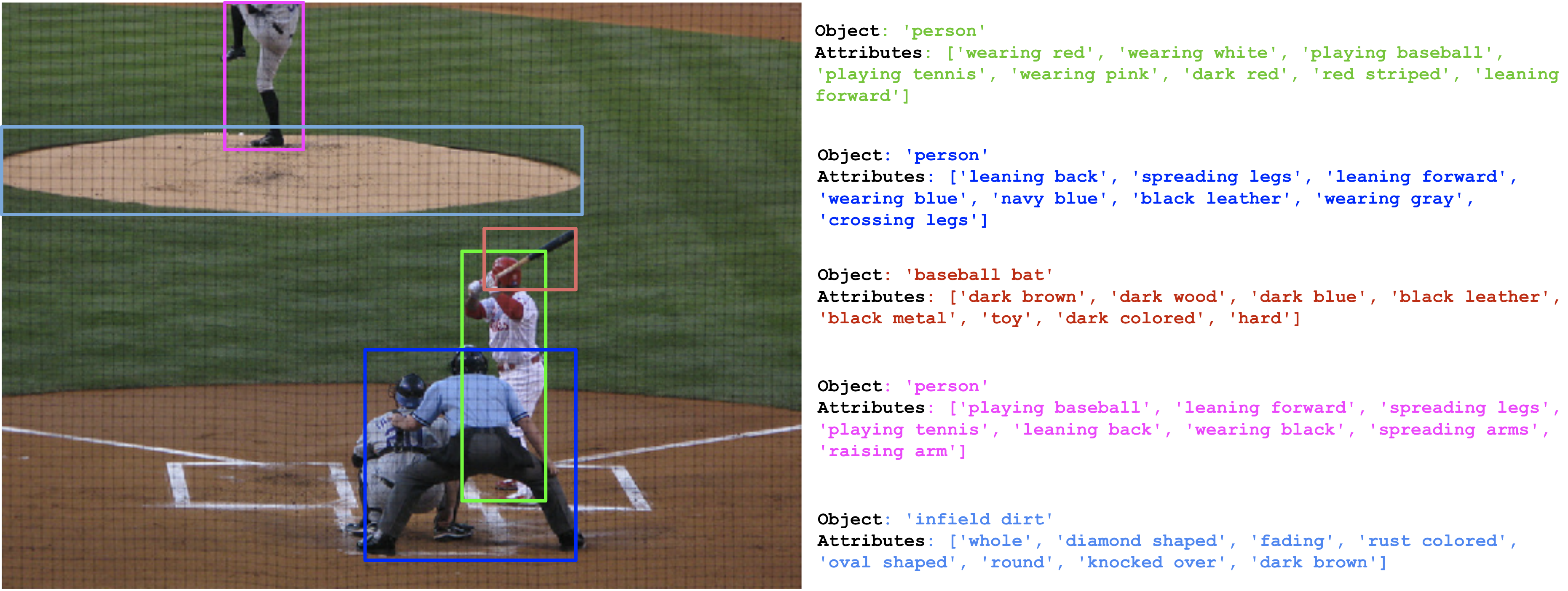}
\caption{Attribute recognition on people playing baseball with object class name and top-8 attributes predicted by LOWA.~\label{example1} The boxes and the predictions are assigned the same color.}
\end{figure*}

\begin{figure*}[tp]
\centering
\includegraphics[width=\textwidth]{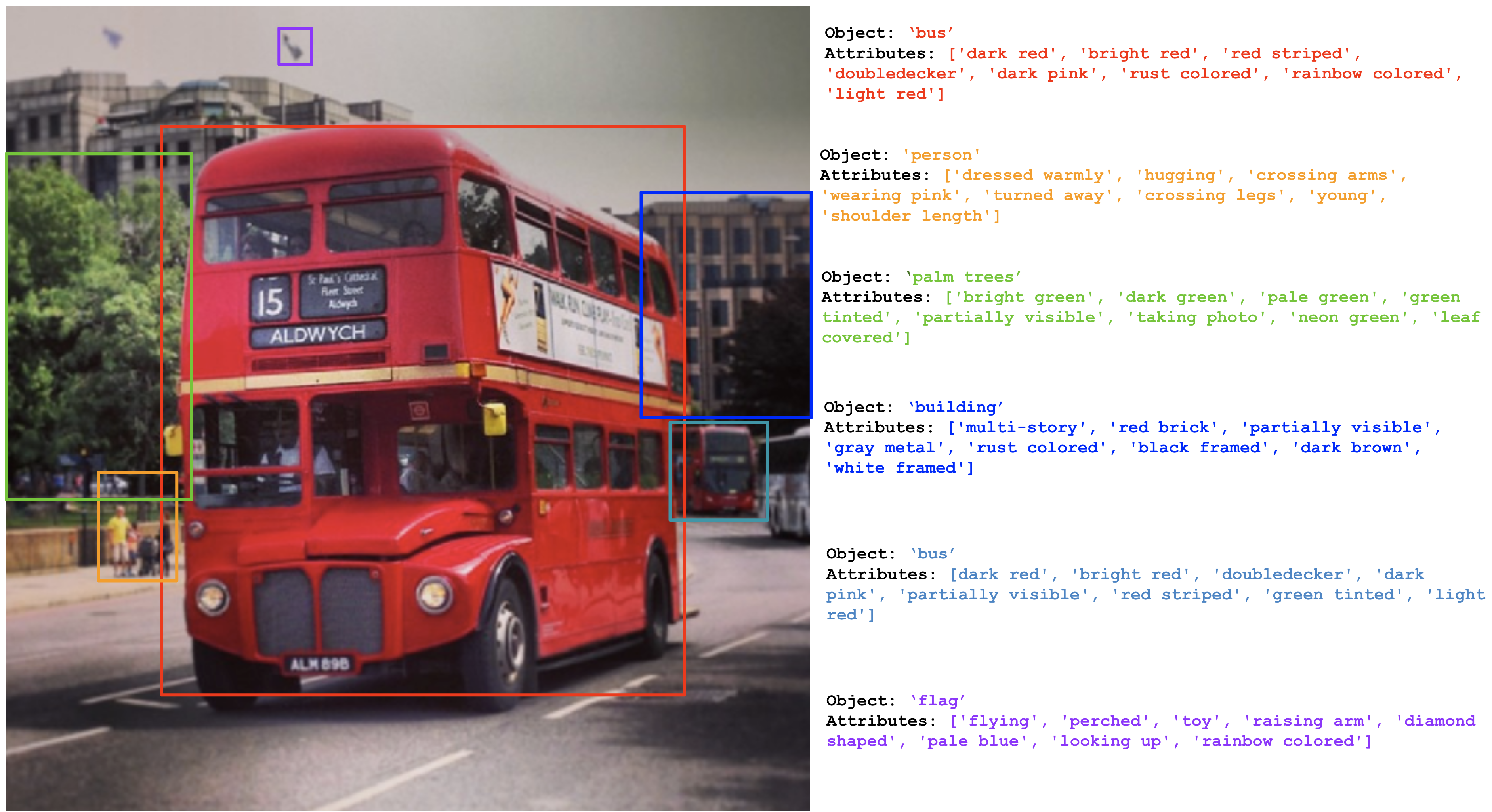}
\caption{Attribute recognition on bus in London with object class name and top-8 attributes predicted by LOWA.~\label{example3}  The boxes and the predictions are assigned the same color. }
\end{figure*}

\begin{figure*}[tp]
\centering
\includegraphics[width=\textwidth]{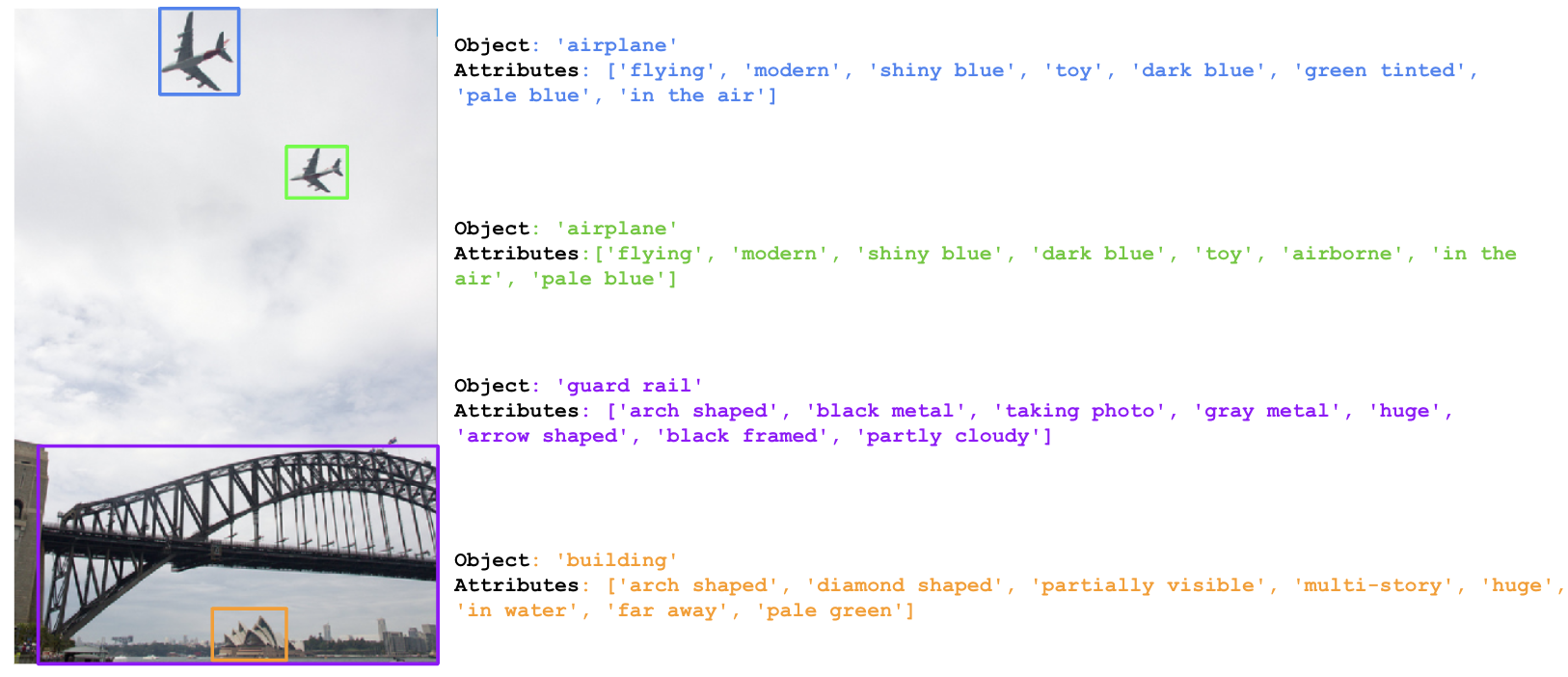}
\caption{Attribute recognition on bridge in Sydney with object class name and top-8 attributes predicted by LOWA.~\label{example2}  The boxes and the predictions are assigned the same color.}
\end{figure*}

\begin{figure*}[tp]
\centering
\includegraphics[width=\textwidth]{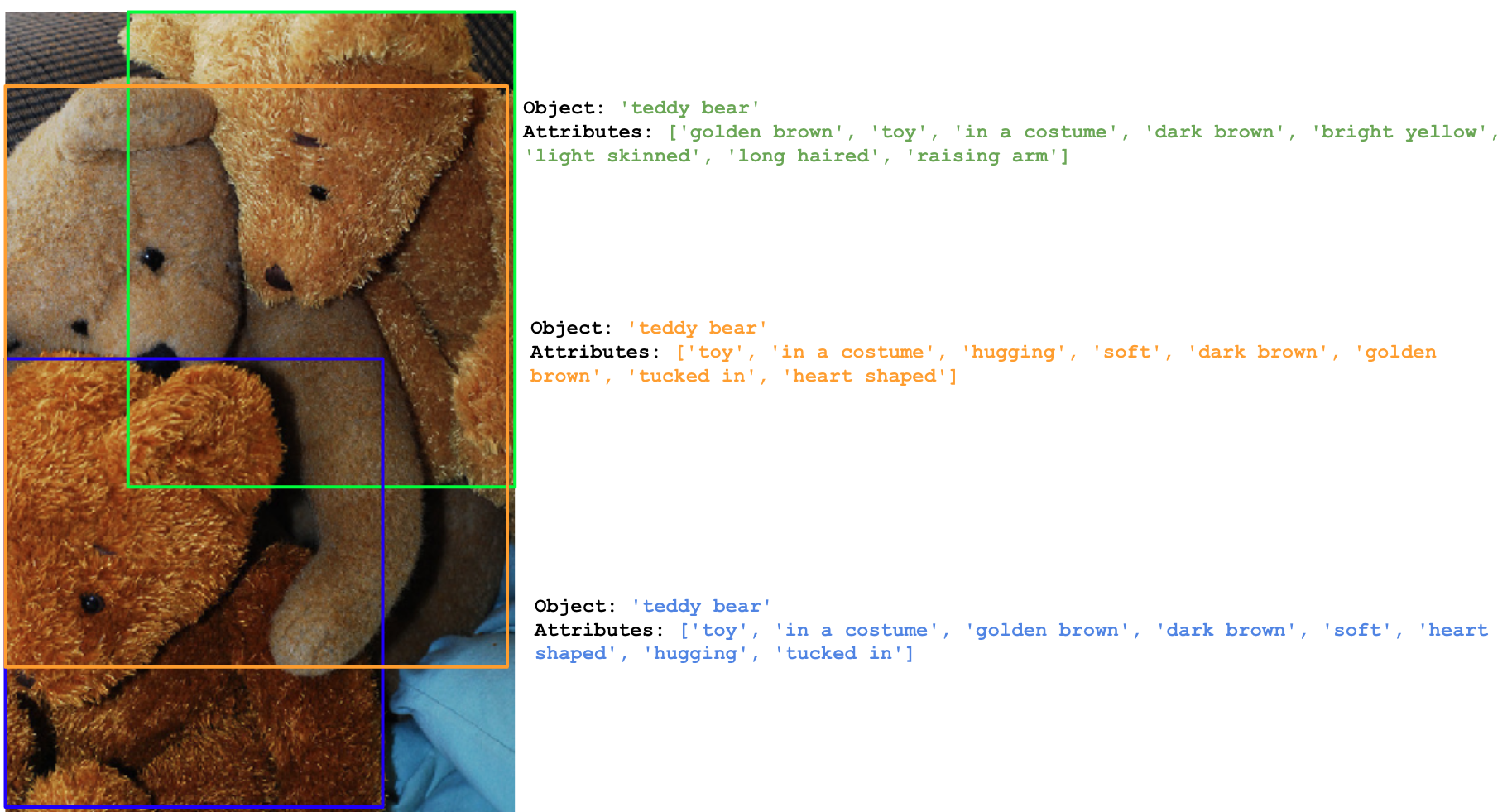}
\caption{Attribute recognition on teddy bear toys with object class name and top-8 attributes predicted by LOWA.~\label{example4}  The boxes and the predictions are assigned the same color.}
\end{figure*}

\begin{figure*}[tp]
\centering
\includegraphics[width=\textwidth]{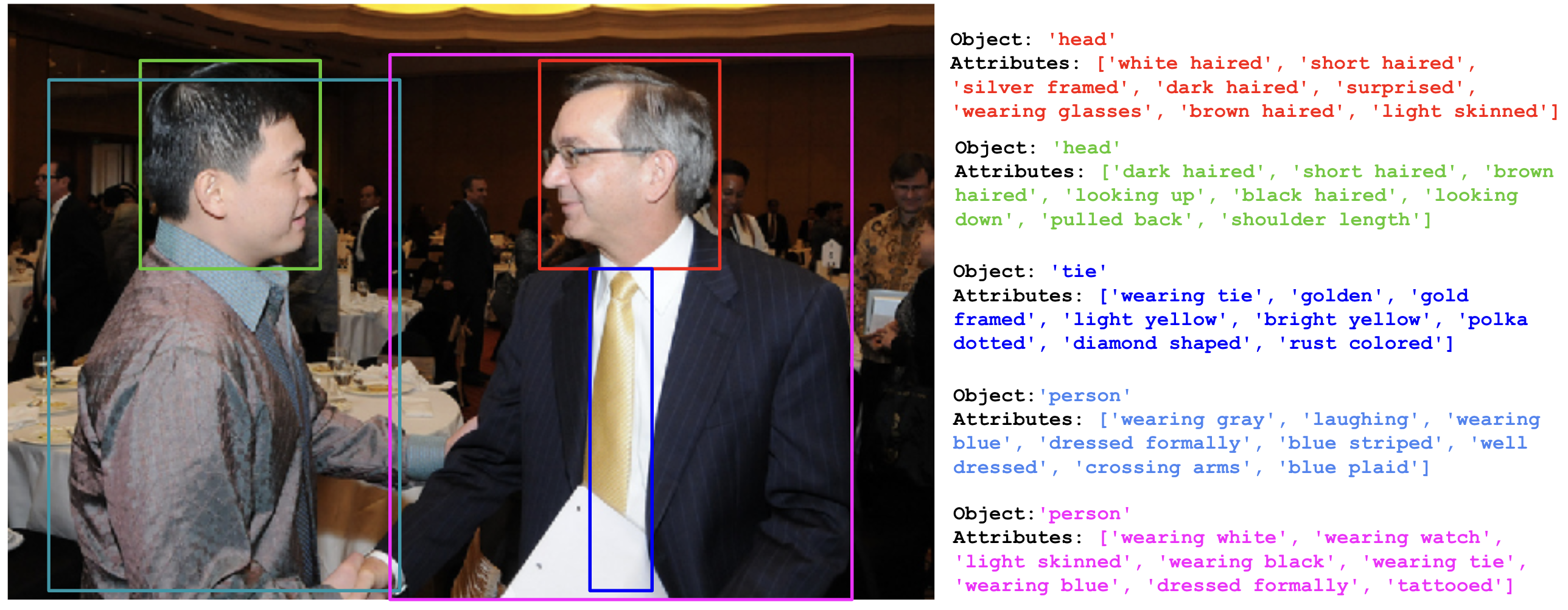}
\caption{Attribute recognition on a party with object class name and top-8 attributes predicted by LOWA.~\label{example5}  The boxes and the predictions are assigned the same color.}
\end{figure*}

\begin{figure*}[tp]
\centering
\includegraphics[width=\textwidth]{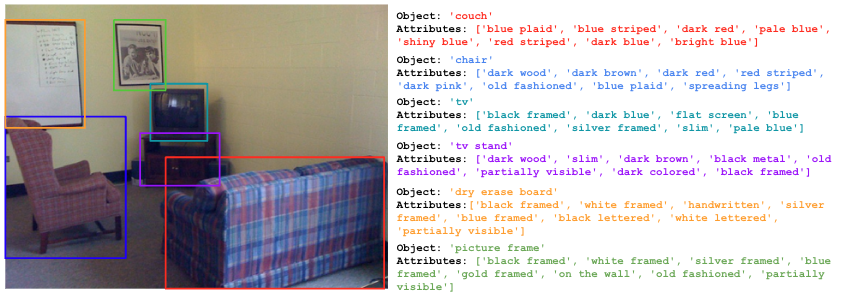}
\caption{Attribute recognition on a room with object class name and top-8 attributes predicted by LOWA.~\label{example6}  The boxes and the predictions are assigned the same color.}
\end{figure*}

In this section, we visualize the results of our model on several example images. First, we verify the accuracy of our \textbf{attribute recognition} results.  We test our model on a variety of objects, including static objects (such as buildings and decorations) and live objects (such as people and animals). We also consider different environments, including indoor and outdoor scenes.  Figures~\ref{example1}, ~\ref{example3}, ~\ref{example2}, ~\ref{example4}, ~\ref{example5}, ~\ref{example6} provide details of the results. For object class name prediction, we use the VAW class names as the input vocabulary and show the top-1 prediction. For attribute recognition, we use the VAW attribute vocabulary as the source text queries and present the top-8 predictions. 

\begin{figure}[tp]
    \centering
     \begin{subfigure}[b]{0.49\textwidth}
         \centering
         \includegraphics[width=\textwidth]{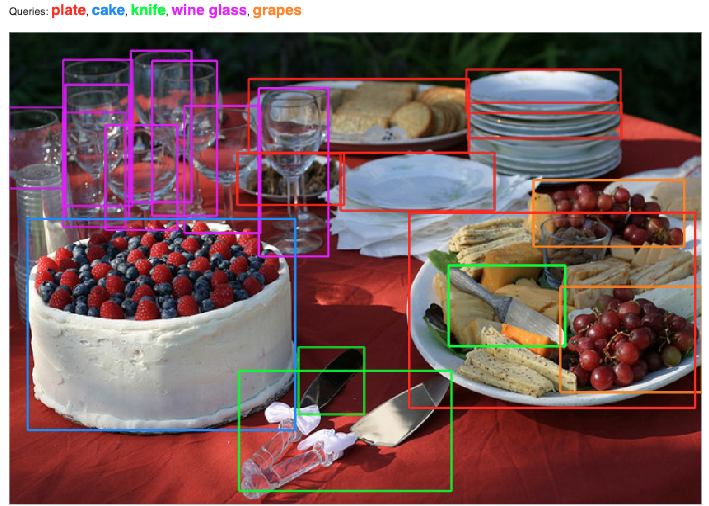}
         \caption{}
         \label{fig:1}
     \end{subfigure}
     \hfill
     \begin{subfigure}[b]{0.48\textwidth}
         \centering
         \includegraphics[width=\textwidth]{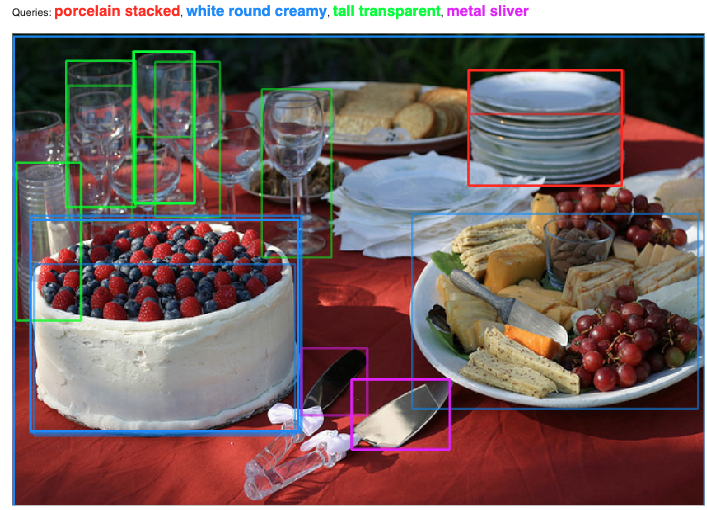}
         \caption{}
         \label{fig:2}
     \end{subfigure}
     \begin{subfigure}[b]{0.49\textwidth}
         \centering
         \includegraphics[width=\textwidth]{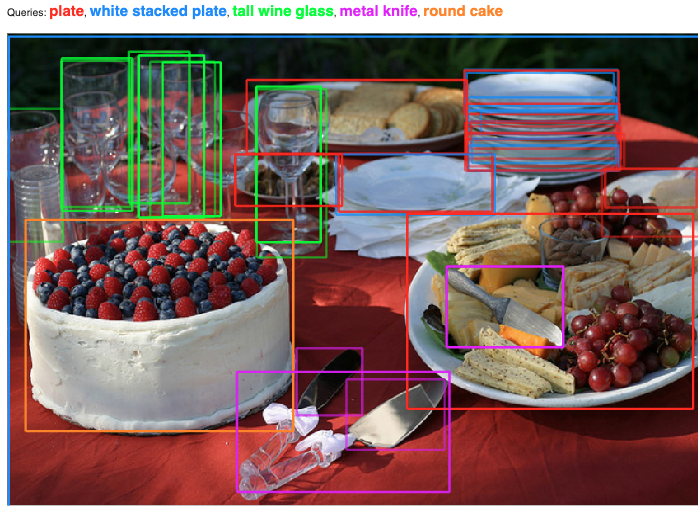}
         \caption{}
         \label{fig:3}
     \end{subfigure}
     \hfill
     \begin{subfigure}[b]{0.48\textwidth}
         \centering
         \includegraphics[width=\textwidth]{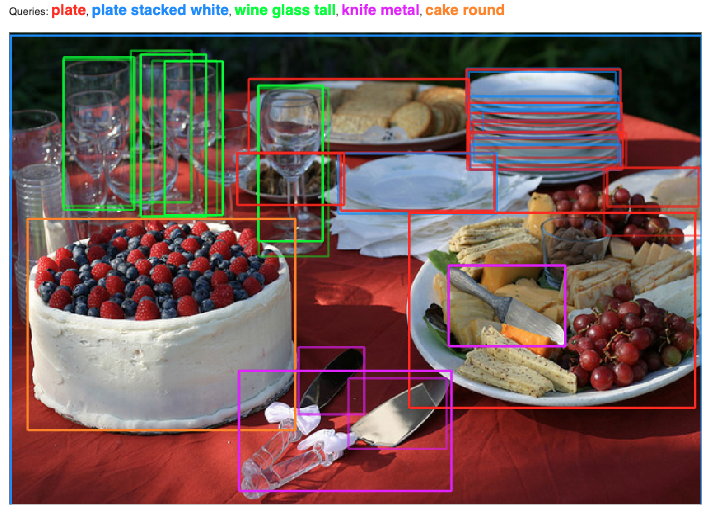}
         \caption{}
         \label{fig:4}
     \end{subfigure}
\caption{Object localization with attributes results of LOWA. Fig.~\ref{fig:1} shows using \textbf{object names} as queries to localize objects; Fig.~\ref{fig:2} illustrates using \textbf{object attributes} as queries to localize objects;  Fig.~\ref{fig:3} is using \textbf{object names + attributes} as queries to localize objects; Fig.~\ref{fig:4} exemplifies \textbf{object attributes + names} as queries to localize objects. Same color means a match between the bounding boxes and the text queries. Strong color quantity means higher confidence/probability scores.}
        \label{fig:fourexample}
\end{figure}

Then, we showcase the \textbf{attribute localization} results in Fig.~\ref{fig:fourexample}. Because our model can take object names, object attributes, and the composition of object names and attributes as queries to localize the target objects. We tested each of these functionalities separately. Figure~\ref{fig:1} shows the results of using object names only; and Figure~\ref{fig:2} shows the results of using attributes only; Figure~\ref{fig:3} shows the results of using object names and attributes in the order of object name + attribute and Fig.~\ref{fig:4} shows the results of using object names and attributes in the order of attribute + object. And it can be observed that our model can take more than one attribute to localize target objects, even though our model is trained with a single attribute mode. For better visualization, each query has a text color for it, and the corresponding detections are shown accordingly with the same color. And the color quantity of the bounding boxes is decided based on their confidence scores. 

\begin{figure*}[tp]
\centering
\includegraphics[width=\textwidth]{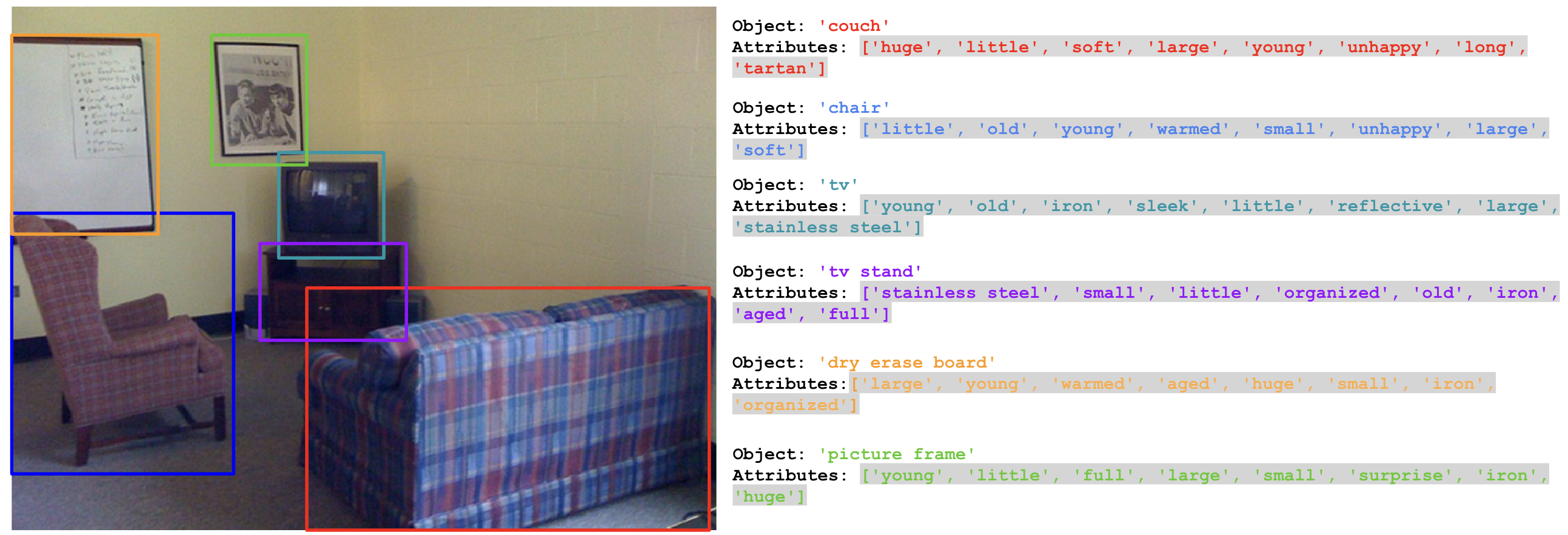}
\caption{A comparison of LOWA predictions with Fig.~\ref{example6}. Fig.~\ref{example6} uses VAW attribute vocabulary while this example uses OVAD attribute (predictions marked in gray) vocabulary.~\label{example11}}
\end{figure*}

LOWA aims to localize objects in the wild. Thus, the input text queries may affect the predictions. We tried both VAW attribute vocabulary shown in Fig.~\ref{example6} and OVAD attribute vocabulary and show the results in Fig.~\ref{example11}. It can be found that with OVAD attribute lists, the predictions sometimes can be irrelevant to the specific object. For example, the $tv$ should not be $young,old$, and the $couch$ should not be both $huge$ and $little$. In contrast, we can identify $tv$ as $black framed$ and the $couch$ as $blue plaid$ with VAW attribute vocabulary. This difference is likely due to the diversity of the vocabularies.  The OVAD dataset only has 117 attributes while VAW has 620 attributes. With a more diverse vocabulary, LOWA is better able to identify the most relevant attributes and make more accurate predictions.

\begin{figure*}[tp]
\centering
\includegraphics[width=\textwidth]{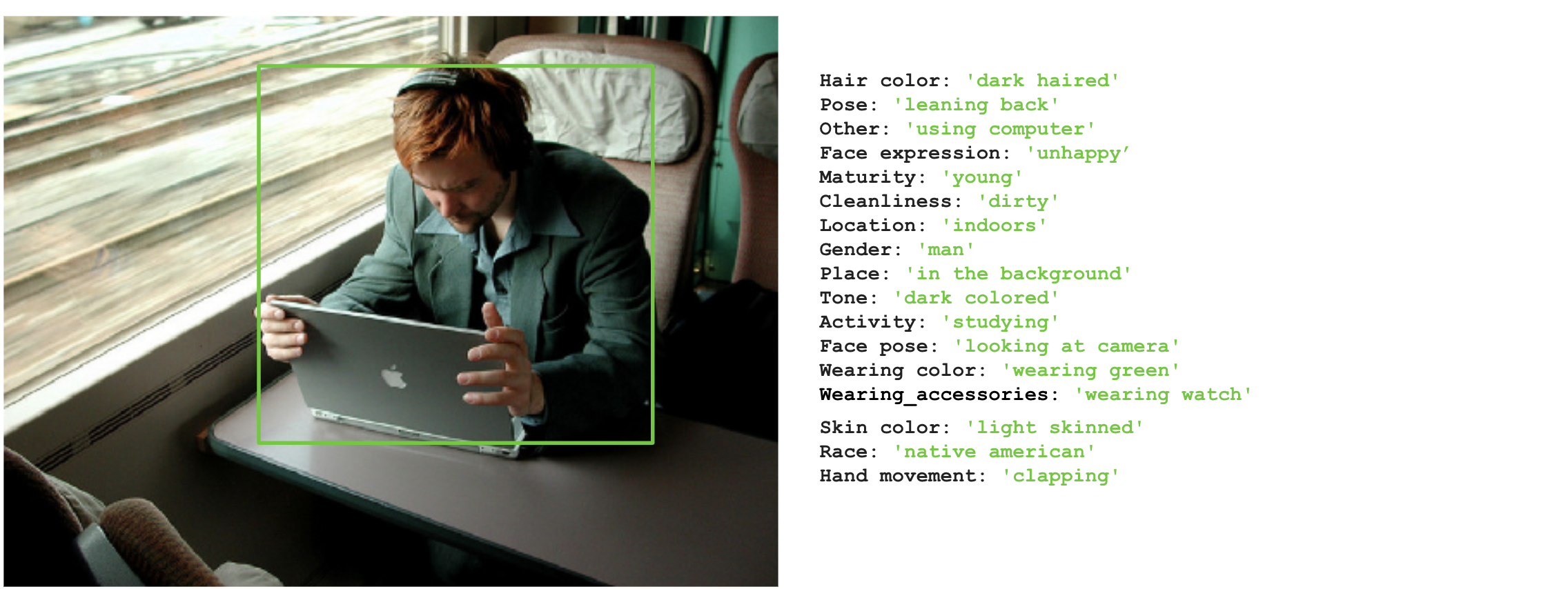}
\caption{An example of LOWA's attribute recognition considering specific attribute categories.~\label{example12}}
\end{figure*}

\begin{figure*}[tp]
\centering
\includegraphics[width=\textwidth]{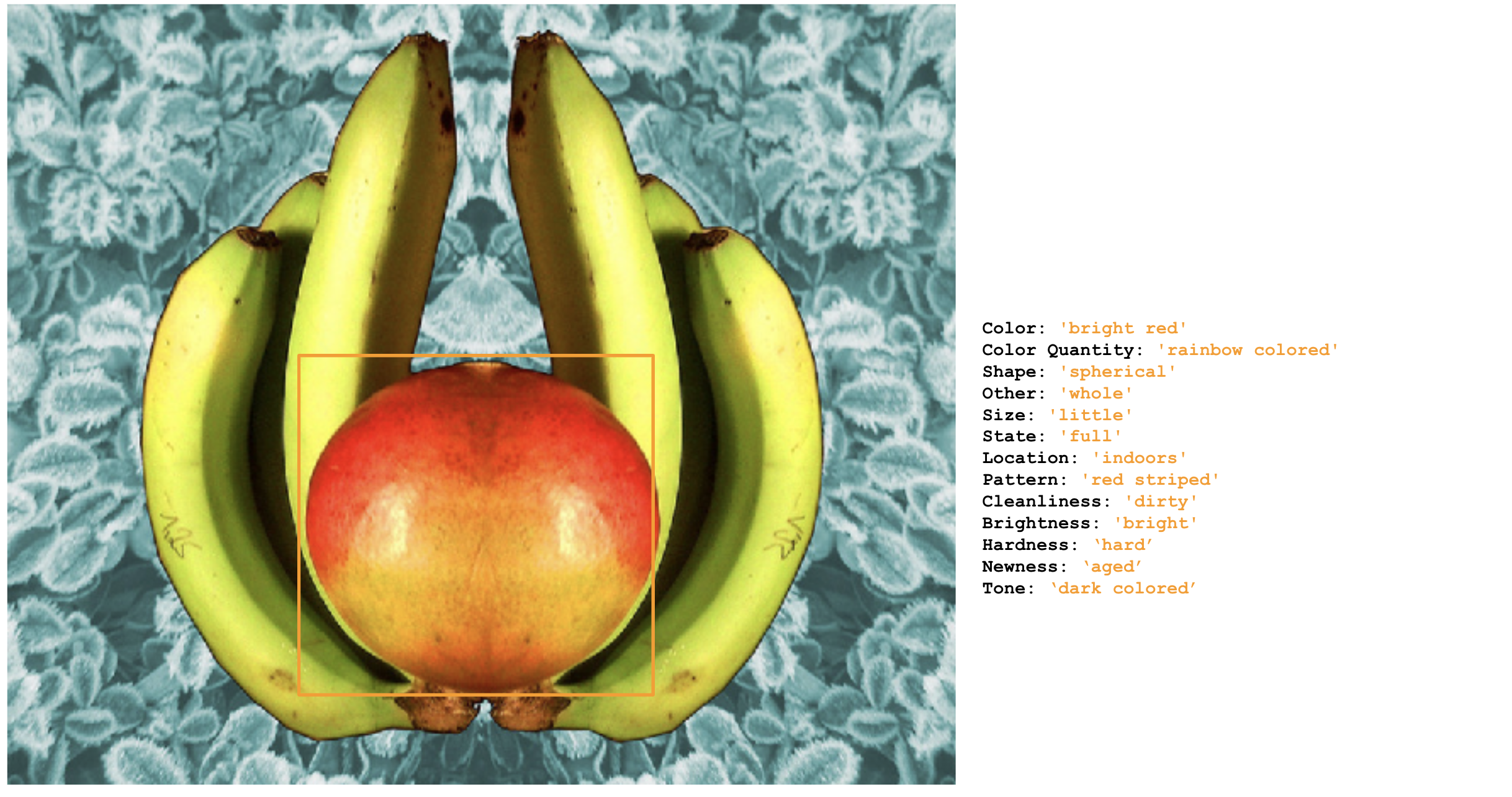}
\caption{One more example of LOWA's attribute recognition considering specific attribute categories.~\label{example13}}
\end{figure*}

Furthermore, we evaluate LOWA's attribute predictions regarding a specific attribute category like color, shape, texture, and so on. For this purpose, we use the attribute lists defined in VAW subcategory. We showcase two examples, Fig.~\ref{example12} is with $person$ and Fig.~\ref{example13} is with $fruit$. We only present the relevant attribute categories based on the object.

\begin{figure*}[tp]
\centering
\includegraphics[width=\textwidth]{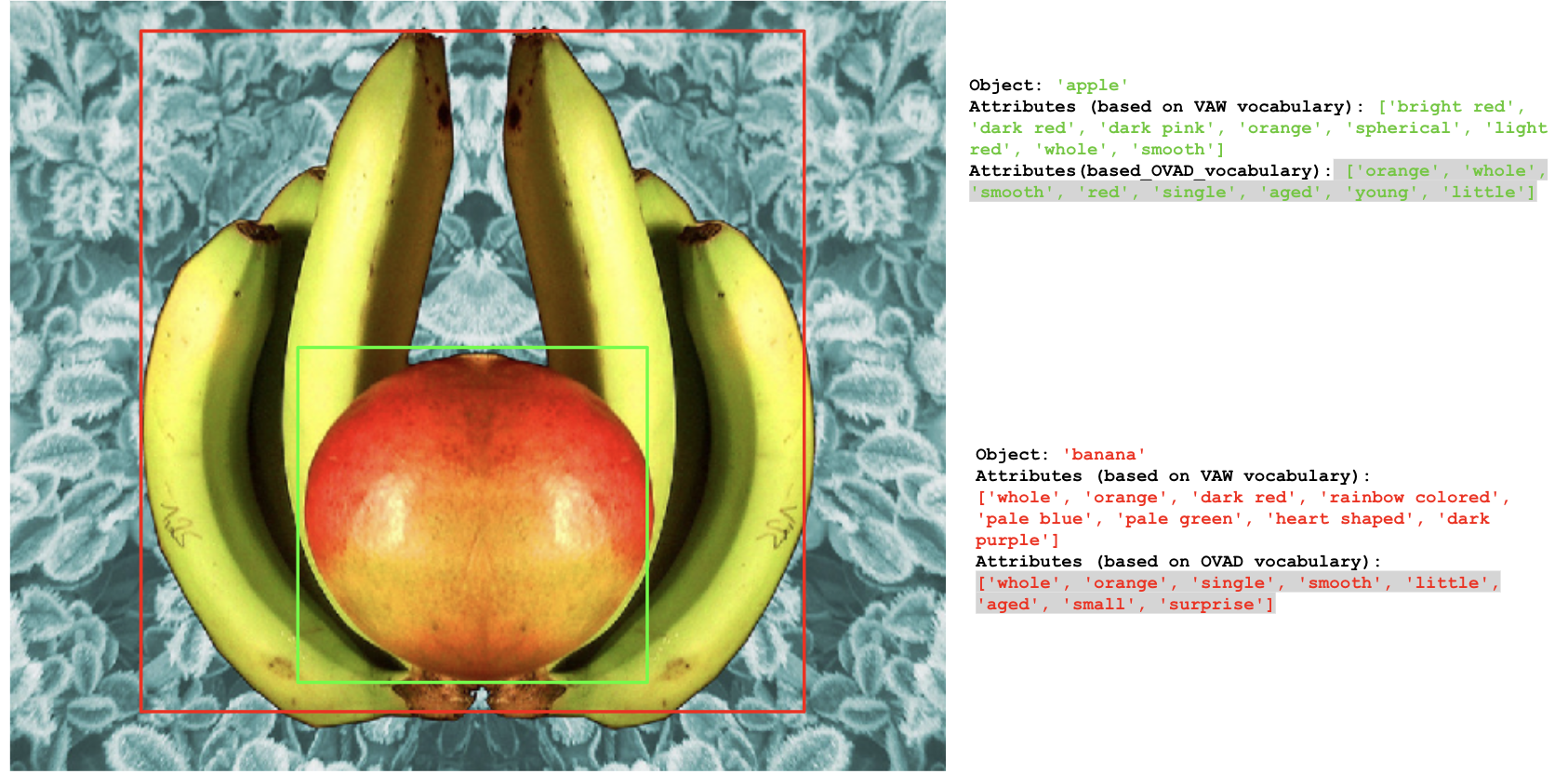}
\caption{LOWA's failure example of attribute recognition on overlapped objects, especially for the $banana$ in the picture. Predictions without gray background is tested with VAW attribute vocabulary while the one with gray background uses OVAD attribute vocabulary.~\label{example14}}
\end{figure*}

LOWA can sometimes make inaccurate predictions, such as identifying the $banana$ in Fig.~\ref{example14} as $dark$ $red$, $pale$ $green$ with VAW vocabulary and $orange$ with OVAD vocabulary. This is likely due to occlusion and overlapping, which can make it difficult for LOWA to identify the object's true color. This limitation could be resolved by adding mask information to help LOWA focus on the object itself in the future.

\end{document}